\documentclass[runningheads]{llncs}

\usepackage{eccv}

\usepackage{eccvabbrv}

\usepackage{graphicx}
\usepackage{booktabs}
\usepackage{mathbbol}
\usepackage[ruled,linesnumbered]{algorithm2e}

\usepackage[accsupp]{axessibility}  

\usepackage{hyperref}

\usepackage{orcidlink}

\begin{document}

\title{Image-Feature Weak-to-Strong Consistency: An Enhanced Paradigm for Semi-Supervised Learning} 

\titlerunning{Image-Feature Weak-to-Strong Consistency}

\author{Zhiyu Wu\orcidlink{0000-0003-2647-408X} \and
Jinshi Cui\orcidlink{0000-0002-8063-4024}}

\authorrunning{Wu and Cui}

\institute{
National Key Laboratory of General Artificial Intelligence, \\
School of Intelligence Science and Technology, Peking University\\
\email{wuzhiyu@pku.edu.cn, cjs@cis.pku.edu.cn}
}

\maketitle

\begin{abstract}
Image-level weak-to-strong consistency serves as the predominant paradigm in semi-supervised learning~(SSL) due to its simplicity and impressive performance. Nonetheless, this approach confines all perturbations to the image level and suffers from the excessive presence of naive samples, thus necessitating further improvement. In this paper, we introduce feature-level perturbation with varying intensities and forms to expand the augmentation space, establishing the image-feature weak-to-strong consistency paradigm. Furthermore, our paradigm develops a triple-branch structure, which facilitates interactions between both types of perturbations within one branch to boost their synergy. Additionally, we present a confidence-based identification strategy to distinguish between naive and challenging samples, thus introducing additional challenges exclusively for naive samples. Notably, our paradigm can seamlessly integrate with existing SSL methods. We apply the proposed paradigm to several representative algorithms and conduct experiments on multiple benchmarks, including both balanced and imbalanced distributions for labeled samples. The results demonstrate a significant enhancement in the performance of existing SSL algorithms.


\end{abstract}
\section{Introduction}
\label{sec:intro}

\begin{figure}[t]
    \centering
    \includegraphics[width=\textwidth]{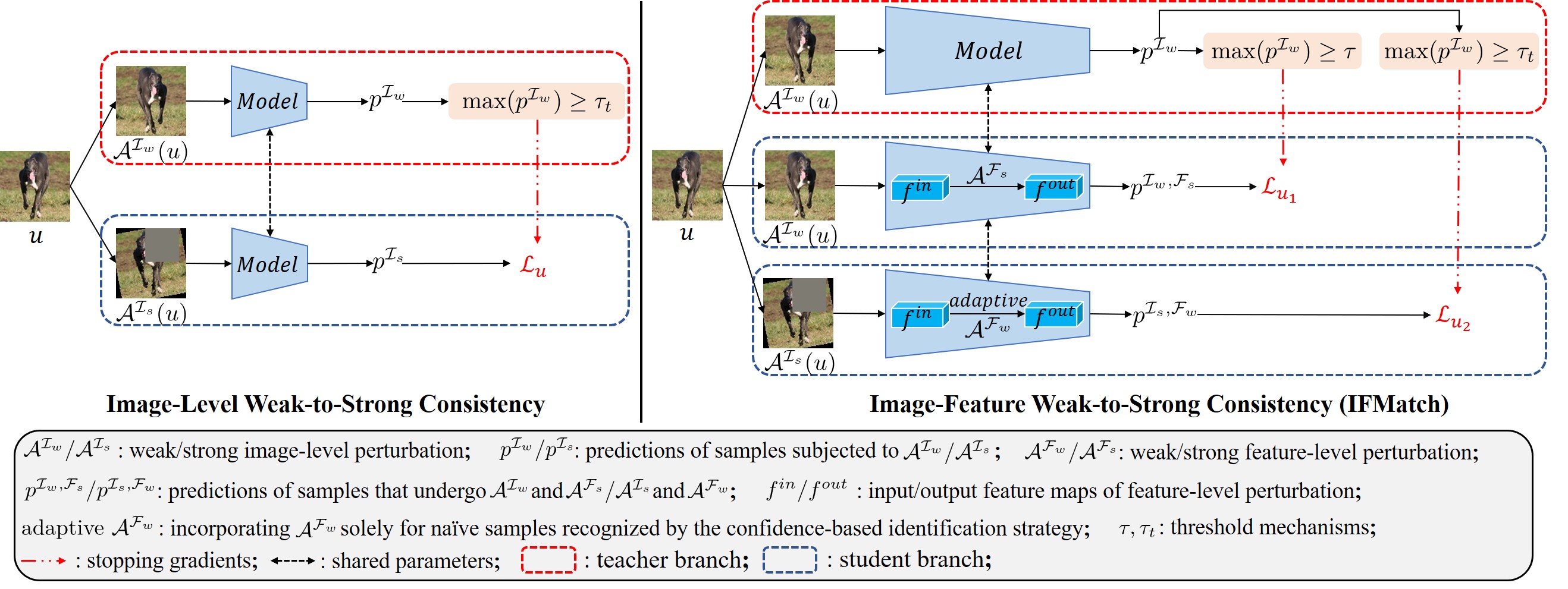}
    \caption{Overview of the old~(left) and proposed~(right) paradigms. Our paradigm introduces feature-level perturbation to expand the augmentation space and facilitates direct interactions between both types of perturbations to boost their synergy.}
    \label{fig:paradigm}
\end{figure}

Semi-supervised learning~(SSL)~\cite{zhu2005semi, rosenberg2005semi, berthelot2019mixmatch, sohn2020fixmatch}, a research topic that aims to reduce manual labeling costs by capitalizing on unlabeled data, has received considerable attention in recent years. Among the proposed methods, the image-level weak-to-strong consistency paradigm introduced by FixMatch~\cite{sohn2020fixmatch} has proven highly effective in leveraging the potential of unlabeled data. Specifically, FixMatch applies varying intensities of image-level perturbations to raw samples and encourages consistent predictions between corresponding weakly and strongly augmented views. Subsequent studies have refined this paradigm primarily in two aspects: (1) improving the utilization ratio of unlabeled data by replacing the constant threshold in FixMatch with a dynamic threshold~\cite{zhang2021flexmatch, wang2022freematch} and (2) facilitating balanced predictions across different classes via refining pseudo-labels based on historical predictions~\cite{berthelot2019remixmatch, chen2023softmatch}. Extensive experiments have demonstrated the effectiveness of these methods in enhancing the image-level weak-to-strong consistency paradigm.

Despite the success of the established paradigm, two challenges persist. Firstly, the existing paradigm confines all perturbations to the image level, thus impeding the exploration of a broader augmentation space and consistency at non-image levels. Secondly, as revealed in~\cite{gui2023enhancing}, a notable proportion of samples, even undergoing strong image-level perturbation, continues to be accurately classified with high confidence, resulting in a loss close to zero. These instances (some examples are shown in \cref{fig:vis}), referred to as \textit{naive samples}, are well-learned and thus fail to boost the model's performance. Therefore, relying solely on image-level perturbation proves insufficient to fully exploit the potential of unlabeled data.

To address the aforementioned challenges, we introduce a novel SSL paradigm named \textbf{I}mage-\textbf{F}eature Weak-to-Strong Consistency~(IFMatch). An overview of the conventional and proposed approaches is shown in \cref{fig:paradigm}. The primary innovation of our method lies in feature-level perturbation, which is motivated by earlier findings that regularizers effective in the input space can be extended to hidden representations~\cite{hinton2012improving, ioffe2015batch, verma2019manifold}. Specifically, feature-level perturbation randomly alters intermediate features in the backbone, establishing the basis for feature-level consistency regulation. Furthermore, to achieve effective feature-level perturbation, we present refined designs focusing on position and strategy. For the perturbation position, we identify two positions to induce weak and strong feature-level perturbations, respectively. Regarding the perturbation strategy, we develop a series of approaches from three perspectives~(`movement', `dropout', and `value'), thereby comprehensively exploring the feature augmentation space. Note that the proposed feature-level perturbation is sample-agnostic and thus differs from previous class-based feature augmentation techniques~\cite{kuo2020featmatch, wang2021regularizing}.

Building upon well-crafted augmentation techniques, we introduce the integration of two types of perturbations as another innovation. Our preliminary investigation~(\cref{tab:ablation_structure}) reveals that a direct fusion of strong image-level and feature-level augmentation engenders destructive perturbations. To mitigate this issue, we propose a triple-branch structure to combine these two perturbation forms in a milder manner, as depicted in \cref{fig:paradigm}. Specifically, the teacher branch, mirroring its counterpart in the existing paradigm, employs predictions from samples subjected to weak image-level perturbation as pseudo-labels. Moreover, the first student branch integrates weak image-level perturbation and strong feature-level perturbation to thoroughly explore the feature augmentation space. Given limited prior research on feature-level perturbation, we delve deeper into the optimal threshold for this branch, revealing a preference for a high constant threshold over a dynamic one. In the second student branch, we combine weak feature-level perturbation with strong image-level perturbation, posing additional challenges for recognizing naive samples. However, the fused perturbation proves excessively arduous for classifying hard samples. Accordingly, we devise a confidence-based identification strategy to dynamically distinguish between naive and challenging samples, thereby introducing weak feature-level perturbation exclusively for naive samples. As the second branch predominantly relies on image-level perturbation, it can seamlessly integrate with existing threshold mechanisms. Prior to our study, UniMatch~\cite{yang2023revisiting} uses feature-level perturbation and a multi-branch structure in semi-supervised semantic segmentation. However, their method is limited to the channel-wise dropout strategy and fails to achieve interactions between the two types of perturbations within a single branch, setting it apart from our paradigm. In summary, our contributions can be outlined as follows.

(1) We introduce the image-feature weak-to-strong consistency paradigm for semi-supervised learning, where the feature-level perturbation and the integration of two types of perturbation within each branch serve as the fundamental enhancement compared to previous methods.

(2) We meticulously design the position and strategy of feature-level perturbation to diversify its intensity and form.

(3) We devise a confidence-based identification strategy to discriminate between naive and challenging samples, thereby introducing extra challenges solely for naive samples in the second student branch.

(4) We conduct experiments on multiple benchmarks, including both balanced and imbalanced distributions for labeled samples. The results demonstrate significant promotion of existing algorithms when following our paradigm.
\section{Related Work}
\label{sec:related_work}

Consistency regulation~\cite{bachman2014learning} represents a fundamental approach to leveraging the potential of unlabeled data. Its core objective is to ensure consistent predictions across diverse perturbed views of the same sample. Various perturbation methods have been explored, such as stochastic augmentation~\cite{laine2016temporal, sajjadi2016regularization}, adversarial attack~\cite{miyato2018virtual}, and mixup~\cite{zhang2017mixup}. Additionally, FixMatch~\cite{sohn2020fixmatch} applies strong data augmentation techniques~\cite{cubuk2020randaugment} to raw images, establishing the image-level weak-to-strong consistency paradigm. This paradigm significantly streamlines the framework and marks a pivotal milestone in semi-supervised learning.

Follow-up studies refine FixMatch along two dimensions: the utilization ratio of unlabeled data and fairness considerations. Regarding the former, numerous dynamic threshold strategies have been proposed. For example, FlexMatch~\cite{zhang2021flexmatch} maps the predefined threshold to class-specific thresholds based on class-wise learning status. SoftMatch~\cite{chen2023softmatch} models sample weights by a dynamic Gaussian function, maintaining a soft margin between unlabeled samples of different confidence levels. FreeMatch~\cite{wang2022freematch} utilizes the confidence on unlabeled data as the adaptive threshold. Moreover, the concept of fairness stems from entropy-based regulation~\cite{krause2010discriminative, arazo2020pseudo, zhao2022dc, grandvalet2004semi, kim2020distribution}. The most representative strategy in this context, Distribution Alignment~\cite{berthelot2019remixmatch}, encourages equal-frequency predictions across classes by refining pseudo-labels based on overall predictions within the unlabeled set.

Concurrently, several studies express similar concerns to the old paradigm. FeatMatch~\cite{kuo2020featmatch} and ISDA~\cite{wang2021regularizing} prevent the exclusive reliance on image-level perturbation by devising class-based feature augmentation approaches. In contrast, our proposed feature-level perturbation is sample-agnostic, circumventing harmful perturbations derived from erroneous pseudo-labels. Additionally, UniMatch~\cite{yang2023revisiting} uses channel-wise dropout to establish an auxiliary branch for feature-level perturbation. In comparison, we present feature-level perturbations with varying intensities and forms, further combining them with image-level perturbations within each branch to boost their synergy. Besides, \cite{gui2023enhancing} observes excessive naive samples and applies more aggressive image-level augmentation to these samples. To address this issue, we propose a confidence-based identification strategy to effectively distinguish between naive and challenging samples and introduce weak feature-level perturbation exclusively for naive samples.
\section{Methodology}
\label{sec:methodology}

\subsection{Image-Level Weak-to-Strong Consistency}
We begin by reviewing the image-level weak-to-strong consistency paradigm, as depicted in the left part of~\cref{fig:paradigm}. Let $\mathcal{D}_L=\{(x_i, y_i)\}_{i=1}^{N_L}$ and $\mathcal{D}_U=\{u_i\}_{i=1}^{N_U}$ represent the labeled and unlabeled datasets, respectively. Here, $x_i$ and $u_i$ denote the labeled and unlabeled training samples, and $y_i$ refers to the one-hot label for the labeled sample $x_i$. We denote the prediction of sample $x$ as $p(y|x)$. Given a batch of labeled and unlabeled data, the model is optimized with the objective $\mathcal{L} = \mathcal{L}_s + \lambda_u \mathcal{L}_u$. Specifically, $\mathcal{L}_s$ indicates the cross-entropy loss~($\mathcal{H}$) for the labeled batch of size $B_L$.
\begin{align}
\label{formula: labeled}
    \mathcal{L}_s = \frac{1}{B_L}\sum_{i=1}^{B_L} \mathcal{H}(y_i, p(y|x_i))
\end{align}
On the other hand, $\mathcal{L}_u$ signifies the image-level consistency regulation between the prediction of the strongly augmented view $\mathcal{A}^{\mathcal{I}_s}(u)$ and the pseudo-label sourced from the corresponding weakly augmented view $\mathcal{A}^{\mathcal{I}_w}(u)$. To mitigate incorrect pseudo-labels, FixMatch~\cite{sohn2020fixmatch} introduces a constant threshold, and subsequent studies propose various dynamic threshold strategies. Without loss of generality, we denote the threshold as $\tau_t$ and define $\mathcal{L}_u$ as follows.
\begin{align}
    \mathcal{L}_u &= \frac{1}{B_U}\sum_{i=1}^{B_U} \mathbb{1}(\text{max}(p^{\mathcal{I}_w}_i) \ge \tau_t) \mathcal{H}(\hat{p}^{\mathcal{I}_w}_i, p^{\mathcal{I}_s}_i)
\end{align}
Here, $p^{\mathcal{I}_w}_i$ is the abbreviation of $\textit{DA}(p(y|\mathcal{A}^{\mathcal{I}_w}(u_i)))$, where \textit{DA} indicates the distribution alignment strategy~\cite{berthelot2019remixmatch}. $\hat{p}^{\mathcal{I}_w}_i$ denotes the one-hot pseudo-label obtained from \textit{argmax}$(p^{\mathcal{I}_w}_i)$. Moreover, $p^{\mathcal{I}_s}_i$ stands for $p(y|\mathcal{A}^{\mathcal{I}_s}(u_i))$. Lastly, $B_U$ corresponds to the batch size of unlabeled data. 

\begin{figure}[t]
    \centering
    \includegraphics[width=0.65\textwidth]{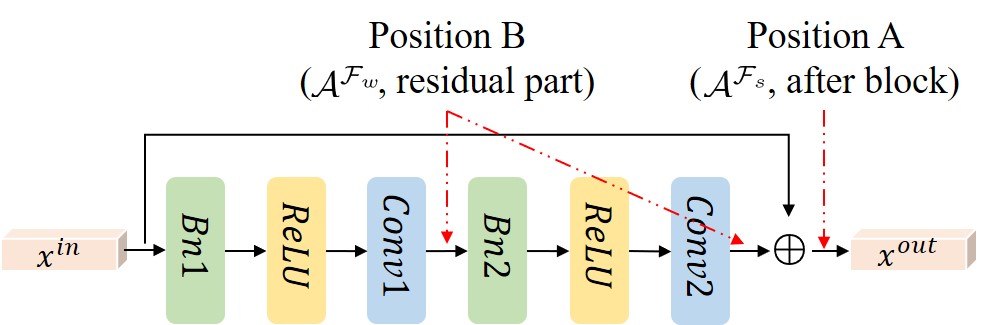}
    \caption{ Candidate positions for introducing feature-level perturbation. Position A perturbs the output of the residual block, representing strong feature-level perturbation $\mathcal{A}^{\mathcal{F}_s}$. Position B perturbs the output of a random convolution within the residual component, corresponding to weak feature-level perturbation $\mathcal{A}^{\mathcal{F}_w}$. $x^{in} / x^{out}$ denotes input/output feature maps of the residual block.}
    \label{fig:position}
\end{figure}

\subsection{Feature-Level Perturbation}
To avoid the exclusive reliance on image-level perturbation, this paper introduces feature-level perturbation varied in intensity and form, thereby expanding the augmentation space. In the following parts, we will elaborate on the implementation of feature-level perturbation regarding its position and strategy.

\textbf{Position refers to where the feature-level perturbation is introduced.} Given that WideResNet~\cite{zagoruyko2016wide} is the prevailing backbone in SSL, we employ its basic component, the residual block~\cite{he2016deep} with pre-activation, as an illustrative example. Building upon previous studies that exploit feature-level regularizers directly after the convolution~\cite{hinton2012improving, ioffe2015batch, verma2019manifold}, we identify two positions for introducing feature-level perturbation, as depicted in \cref{fig:position}. In position A, the method perturbs the summation of the residual and identity connections, \ie, the output of the residual block. As position A acts as the bottleneck for feedforward, perturbation at position A exhibits high intensity and is therefore regarded as strong feature-level perturbation $\mathcal{A}^{\mathcal{F}_s}$. Additionally, position B perturbs the output of a random convolution within the residual component, which is relatively milder than position A. Therefore, perturbation at position B is considered as weak feature-level perturbation $\mathcal{A}^{\mathcal{F}_w}$. It is worth noting that both position choices are applicable to all residual blocks, and the algorithm randomly selects one position at each iteration to insert feature-level perturbation.

\textbf{Strategy denotes the operation of feature-level perturbation.} We first review strong image-level perturbation strategies $\mathcal{A}^{\mathcal{I}_s}$~\cite{cubuk2020randaugment}, which can be roughly divided into three groups: `movement', `dropout', and `value'. Specifically, `movement' comprises spatial perturbations like translation and shearing, `dropout' corresponds to the Cutout technique~\cite{devries2017improved}, and `value' represents operations that modify pixel values while preserving pixel relationships, such as adjusting contrast, brightness, and histogram. Considering the effectiveness of $\mathcal{A}^{\mathcal{I}_s}$, we develop feature-level perturbation strategies from the same perspectives. For `movement', we implement feature map translation and shearing along the X-axis and Y-axis. Regarding `dropout', we apply channel-wise and spatial-wise dropout on hidden representations. Concerning `value', we employ random-sized convolutions to achieve local smoothing, subsequently using the weighted sum of smoothed and input feature maps as the output. During training, the model randomly selects one candidate strategy at each iteration to perturb hidden representations at the selected position. \textit{For the detailed implementation of feature-level perturbation strategies, please refer to Appendix Sec.1.}

\subsection{Image-Feature Weak-to-Strong Consistency}
\label{sec: new paradigm}
Building upon well-designed image-level and feature-level perturbations, we can integrate them to achieve consistency regulation in both dimensions. Intuitively, a trivial approach is combining strong image-level and feature-level perturbations in one branch to construct a dual-branch structure similar to the old paradigm. However, such a toy paradigm engenders destructive perturbations according to our pilot study~(\cref{tab:ablation_structure}). Accordingly, we integrate these two forms of perturbations in a more moderate manner, developing the image-feature weak-to-strong consistency paradigm~(IFMatch) depicted in the right part of \cref{fig:paradigm}.

Our paradigm retains the traditional teacher branch, utilizing predictions $\hat{p}^{\mathcal{I}_w}$ of samples subjected to weak image-level perturbation as pseudo-labels. Furthermore, we introduce two student branches to integrate both types of perturbations. Mathematically, given a batch of labeled and unlabeled data, the model is optimized using the objective $\mathcal{L} = \mathcal{L}_s + \lambda_u (\mathcal{L}_{u_1} + \mathcal{L}_{u_2})$, where $\mathcal{L}_s$ denotes the cross-entropy loss for labeled data, and $\mathcal{L}_{u_1}$ and $\mathcal{L}_{u_2}$ represent the consistency regulation for unlabeled data in the two student branches respectively.

In the first student branch, we integrate weak image-level perturbation $\mathcal{A}^{\mathcal{I}_w}$ and strong feature-level perturbation $\mathcal{A}^{\mathcal{F}_s}$ to comprehensively explore the feature perturbation space. \textit{Furthermore, we employ the constant threshold $\tau$ adopted in FixMatch to filter out incorrect pseudo-labels.} This choice arises from the observation that existing dynamic threshold mechanisms are less suitable for $\mathcal{A}^{\mathcal{F}_s}$, especially in challenging tasks. Thus, the unsupervised loss $\mathcal{L}_{u_1}$ in the first student branch can be expressed as follows.
\begin{align}
\label{formula: lu1}
    \mathcal{L}_{u_1} = \frac{1}{B_U}\sum_{i=1}^{B_U} \mathbb{1}(\text{max}(p^{\mathcal{I}_w}_i) \ge \tau) \mathcal{H}(\hat{p}^{\mathcal{I}_w}_i, p^{\mathcal{I}_w,\mathcal{F}_s}_i)
\end{align}
Here, $p^{\mathcal{I}_w,\mathcal{F}_s}_i$ denotes the prediction of $u_i$ in the first branch.

\begin{algorithm}[t]
    \caption{Image-Feature Weak-to-Strong Consistency~(IFMatch)}
    \label{algorithm}
    \KwIn{labeled data $\mathcal{D}_L=\{(x_i, y_i)\}_{i=1}^{N_L}$, unlabeled data $\mathcal{D}_U=\{u_i\}_{i=1}^{N_U}$, batch sizes $B_L$ and $B_U$, loss weight $\lambda_u$, total iterations $T$, perturbations $\{\mathcal{A}^{\mathcal{I}_w}$, $\mathcal{A}^{\mathcal{I}_s}$, $\mathcal{A}^{\mathcal{F}_w}$,
    $\mathcal{A}^{\mathcal{F}_s}\}$, constant threshold $\tau$, threshold mechanism $\tau_t$, target confidence $\{h_1, h_2, \cdots, h_{N_U}\}$, mask $\{\mathcal{M}_1, \mathcal{M}_2, \cdots, M_{N_U} \}$}
    \For{$t \leftarrow 1$ \KwTo $T$}{

        \tcp{the same parts as the old paradigm}
        Compute supervised loss $\mathcal{L}_s$, pseudo-labels $p^{\mathcal{I}_w}$ and $\hat{p}^{\mathcal{I}_w}$, and threshold $\tau_t$ \;

        \tcp{the first student branch}
        Compute predictions $p^{\mathcal{I}_w, \mathcal{F}_s}$ and loss $\mathcal{L}_{u_1}$ using \cref{formula: lu1} \;

        \tcp{the second student branch}
        Decide whether to introduce $\mathcal{A}^{\mathcal{F}_w}$ for $u_i$ in the second student branch according to the mask $\mathcal{M}_i$ using \cref{formula: perturbation} and \cref{formula: compare}\;
        
        Compute predictions $p^{\mathcal{I}_s, \mathcal{F}_w}$ and loss $\mathcal{L}_{u_2}$ using \cref{formula: lu2} \;
        
        \tcp{update model}
        Back-propagate loss $\mathcal{L} = \mathcal{L}_s + \lambda_u (\mathcal{L}_{u_1} + \mathcal{L}_{u_2})$ and update parameters \;

        \tcp{Confidence-based identification strategy}
        Record target confidence $h_i = p_{i, j}^{\mathcal{I}_s, \mathcal{F}_w} (j=\hat{p}^{\mathcal{I}_w}_i)$ and mask $\mathcal{M}_i=\mathbb{1}(h_i \ge \tau_t)$\;
    }
\end{algorithm}

In the second student branch, we combine strong image-level perturbation $\mathcal{A}^{\mathcal{I}_s}$ and weak feature-level perturbation $\mathcal{A}^{\mathcal{F}_w}$ to provide additional challenges in recognizing naive samples. However, a direct fusion of $\mathcal{A}^{\mathcal{I}_s}$ and $\mathcal{A}^{\mathcal{F}_w}$ poses obstacles in classifying hard samples. Accordingly, we present a confidence-based identification strategy to distinguish between naive and challenging samples, thus introducing extra challenges exclusively for naive samples, which will be detailed in \cref{sec: adaptive_combination}. Moreover, given the central role of $\mathcal{A}^{\mathcal{I}_s}$, \textit{we leverage existing threshold mechanisms $\tau_t$ to exclude misleading pseudo-labels}, facilitating seamless integration with previous algorithms. Consequently, the unsupervised loss $\mathcal{L}_{u_2}$ in the second branch can be defined as follows.
\begin{align}
\label{formula: lu2}
    \mathcal{L}_{u_2} = \frac{1}{B_U}\sum_{i=1}^{B_U} \mathbb{1}(\text{max}(p^{\mathcal{I}_w}_i) \ge \tau_t) \mathcal{H}(\hat{p}^{\mathcal{I}_w}_i, p^{\mathcal{I}_s,\mathcal{F}_w}_i)
\end{align}
where $p^{\mathcal{I}_s,\mathcal{F}_w}_i$ represents the prediction of sample $u_i$ in the second branch. Particularly, the use of distinct thresholds in the two student branches is motivated not only by experiments~(\cref{tab:threshold_branchI}) but also by an inherent rationale. Specifically, the threshold serves to balance the utilization ratio of unlabeled data and pseudo-label accuracy~\cite{wang2022freematch, chen2023softmatch}. Furthermore, $\mathcal{A}^{\mathcal{F}_s}$ and $\mathcal{A}^{\mathcal{I}_s}$ act as the central driving forces behind the respective student branch. Due to the unique characteristics of $\mathcal{A}^{\mathcal{F}_s}$ and $\mathcal{A}^{\mathcal{I}_s}$, they exhibit varying preferences in the trade-off between the utilization and pseudo-label accuracy, resulting in the different desired threshold mechanisms in the two student branches. For a further understanding of our paradigm, please refer to \cref{algorithm}.

\begin{figure}[t]
    \centering
    \includegraphics[width=0.65\textwidth]{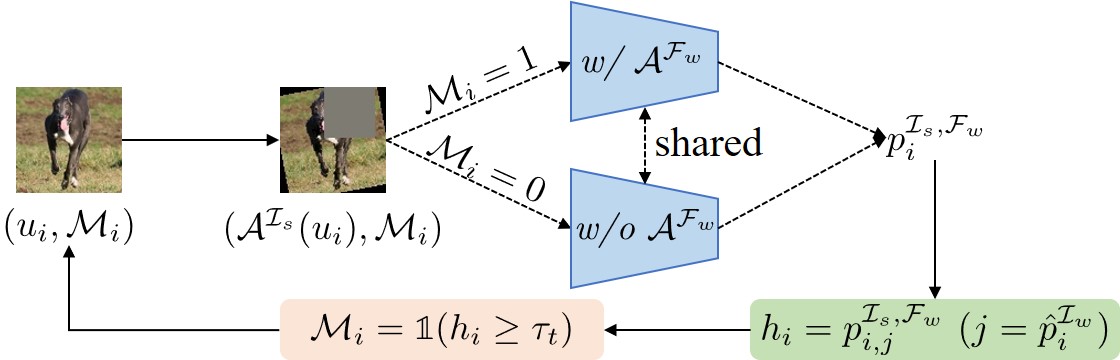}
    \caption{Overview of the confidence-based identification strategy. The approach records sample-wise target confidence in the second student branch and identifies naive samples by comparing the confidence with the threshold in the second student branch. \textit{w/} and \textit{w/o} abbreviates \textit{with} and \textit{without}, respectively.}
    \label{fig:CBI}
\end{figure}

\subsection{Confidence-Based Identification}
\label{sec: adaptive_combination}

As analyzed in~\cref{sec: new paradigm}, the combination of strong image-level perturbation $\mathcal{A}^{\mathcal{I}_s}$ and weak feature-level perturbation $\mathcal{A}^{\mathcal{F}_w}$ poses supplementary challenges for naive samples yet incurs exceeding difficulty for hard ones. Likewise, previous research has encountered similar challenges in expanding the augmentation space. For example, UniMatch~\cite{yang2023revisiting} observes a performance degradation caused by the concatenation of channel-wise dropout and $\mathcal{A}^{\mathcal{I}_s}$ and finally denies the feasibility of combining $\mathcal{A}^\mathcal{I}$ and $\mathcal{A}^\mathcal{F}$ within one branch. Besides, \cite{gui2023enhancing} imposes more aggressive image-level perturbations exclusively on naive samples. To distinguish between naive and challenging samples, their approach records sample-wise loss and utilizes the OTSU~\cite{otsu1979threshold} method for distinction. However, OTSU is conventionally applied to the segmentation of bimodal gray-scale images, whereas the histogram of loss values typically exhibits a descending trend. As shown in \cref{fig:adaptive}, their approach tends to produce identification biased toward naive samples.

To effectively identify the naive samples, we propose a confidence-based identification strategy, which is depicted in \cref{fig:CBI}. Similar to the threshold-based rule for pseudo-label filtering that only predictions exceeding the confidence threshold qualify as pseudo-labels, the proposed strategy can be succinctly concluded as: `\textit{only strongly augmented samples $\mathcal{A}^{\mathcal{I}_s}(u)$ with target confidence exceeding the threshold are considered naive and necessitate additional weak feature-level perturbation $\mathcal{A}^{\mathcal{F}_w}$}'. Specifically, the approach comprises two steps. Firstly, we record the confidence linked to the pseudo-label within the predicted distribution of the second branch. Mathematically, the target confidence $h_i$ for sample $u_i$ in the second student branch can be obtained as follows.
\begin{align}
    h_i = p_{i, j}^{\mathcal{I}_s, \mathcal{F}_w} \  (j=\hat{p}^{\mathcal{I}_w}_i)
\end{align}
It is important to clarify that $p_i^{\mathcal{I}_s, \mathcal{F}_w}$ solely represents the prediction of sample $u_i$ in the second branch and has nothing to do with the introduction of $\mathcal{A}^{\mathcal{F}_w}$ for $u_i$. Secondly, the algorithm generates a binary mask for each sample to determine whether to introduce additional $\mathcal{A}^{\mathcal{F}_w}$ by comparing the target confidence with the threshold $\tau_t$ in the second student branch. Consequently, the mask for sample $u_i$, denoted as $\mathcal{M}_i$, and the perturbations applied to $u_i$ in the second student branch can be expressed as follows.
\begin{align}
\label{formula: perturbation}
    \mathcal{M}_i &= \mathbb{1}(h_i \ge \tau_t) \\
\label{formula: compare}
    \text{Perturbations for } u_i &=
    \begin{cases}
        \{\mathcal{A}^{\mathcal{I}_s}, \mathcal{A}^{\mathcal{F}_w}\} &  \mathcal{M}_i = 1 \\
        \{\mathcal{A}^{\mathcal{I}_s}\} &  \mathcal{M}_i = 0 \\
    \end{cases}
\end{align}
The approach dynamically updates $\mathcal{M}_i$ throughout the training process, thereby considering both the learning status of individual samples and model performance when introducing $\mathcal{A}^{\mathcal{F}_w}$.
\section{Experiments}
\label{sec:experiments}

\begin{table*}[t]
    \scriptsize
    \centering
    \caption{Accuracy~(\%) on four benchmarks with varying numbers of labels. \textbf{Bold} indicates the best performance, and \underline{underline} denotes the second best. We omit the standard deviation of performance, as the proposed paradigm performs stable across different random seeds. IFMatch (*) represents the proposed paradigm using the threshold mechanism of * in the second student branch. In Particular, given that SoftMatch proposes a sample weight strategy rather than a threshold scheme, IFMatch~(Soft) inherits its weight scheme when computing $\mathcal{L}_{u_2}$ and replaces the $\tau_t$ in \cref{formula: perturbation} with the mean value of the dynamic Gaussian in SoftMatch when identifying naive samples.}
    \tabcolsep=0.05cm
    \begin{tabular}{c|ccc|ccc|cc|cc|cc}
    \toprule
    Datasets & \multicolumn{3}{c}{CIFAR-10} \vline & \multicolumn{3}{c}{CIFAR-100} \vline & \multicolumn{2}{c}{SVHN} \vline & \multicolumn{2}{c}{STL-10} \vline & \multicolumn{2}{c}{ImageNet} \\
    \midrule
    \# Label & 40 & 250 & 4000 & 400 & 2500 & 10000 & 40 & 1000 & 40 & 1000 & 100k & 400k \\
    \midrule
    VAT~\cite{miyato2018virtual} & 25.34 & 58.97 & 89.49 & 14.80 & 53.16 & 67.86 & 25.25 & 95.89 & 25.26 & 62.05 & - & -\\
    MeanTeacher~\cite{tarvainen2017mean} & 29.91 & 62.54 & 91.90 & 18.89 & 54.83 & 68.25 & 63.91 & 96.73 & 28.28 & 66.10 & - & -\\
    MixMatch~\cite{berthelot2019mixmatch} & 63.81 & 86.37 & 93.34 & 32.41 & 60.24 & 72.22 & 69.40 & 96.31 & 45.07 & 78.30 & - & -\\
    ReMixMatch~\cite{berthelot2019remixmatch} & 90.12 & 93.70 & 95.16 & 57.25 & 73.97 & 79.98 & 75.96 & 94.84 & 67.88 & 93.26 & - & -\\
    UDA~\cite{xie2020unsupervised} & 89.38 & 94.84 & 95.71 & 53.61 & 72.27 & 77.51 & 94.88 & 98.11 & 62.58 & 93.36 & - & -\\
    Dash~\cite{xu2021dash} & 91.07 & 94.84 & 95.64 & 55.18 & 72.85 & 78.12 & 97.81 & 98.03 & 65.48 & 93.61 & - & - \\
    MPL~\cite{pham2021meta} & 93.38 & 94.24 & 95.45 & 53.74 & 72.29 & 78.26 & 90.67 & 97.72 & 64.24 & 93.34 & - & - \\
    \midrule
    FixMatch~\cite{sohn2020fixmatch} & 92.53 & 95.14 & 95.79 & 53.58 & 71.97 & 77.80 & 96.19 & 98.04 & 64.03 & 93.75 & 56.34 & 67.72 \\
    IFMatch~(Fix) & 95.82 & 95.97 & \textbf{96.49} & 66.26 & 74.65 & \underline{80.01} & 97.75 & 98.15 & 78.54 & \underline{95.28} & 61.26 & \underline{71.48} \\
    \midrule
    FlexMatch~\cite{zhang2021flexmatch} & 95.03 & 95.02 & 95.81 & 60.06 & 73.51 & 78.10 & 91.81 & 93.28 & 70.85 & 94.23 & 58.15 & 68.69 \\
    IFMatch~(Flex) & \textbf{95.90} & \underline{95.99} & 96.42 & 66.28 & \underline{75.03} & 79.87 & 97.76 & 98.04 & 79.83 & 95.24 & \underline{61.75} & 71.14 \\
    \midrule
    SoftMatch~\cite{chen2023softmatch} & 95.09 & 95.18 & 95.96 & 62.90 & 73.34 & 77.97 & 97.67 & 97.99 & 78.58 & 94.27 & 59.43 & 70.51 \\
    IFMatch~(Soft) & \underline{95.84} & 95.93 & 96.36 & \textbf{67.02} & 74.99 & 79.96 & \underline{98.04} & \textbf{98.22} & 82.46 & 95.22 & \textbf{62.44} & \textbf{71.66} \\
    \midrule
    FreeMatch~\cite{wang2022freematch} & 95.10 & 95.12 & 95.90 & 62.02 & 73.53 & 78.32 & 98.03 & 98.04 & \underline{84.44} & 94.37 & 58.48 & 69.48 \\
    IFMatch~(Free) & 95.83 & \textbf{96.02} & \underline{96.44} & \underline{66.65} & \textbf{75.16} & \textbf{80.08} & \textbf{98.08} & \underline{98.18} & \textbf{84.75} & \textbf{95.29} & 61.69 & 71.30 \\
    \bottomrule
    \end{tabular}
    \vspace{-0.4cm}
    \label{tab:performance}
\end{table*}

\subsection{Balanced Semi-Supervised Learning}

\textbf{Settings.} We conduct experiments on multiple benchmarks, including CIFAR-10/100~\cite{krizhevsky2009learning}, SVHN~\cite{netzer2011reading}, STL-10~\cite{coates2011analysis}, and ImageNet~\cite{deng2009imagenet}, with various numbers of labeled samples. In the context of balanced SSL, we maintain a balanced distribution for labeled samples. The choice of backbone architecture adheres to established practices, employing specific models for different datasets: WRN-28-2~\cite{zagoruyko2016wide} for CIFAR-10 and SVHN, WRN-28-8 for CIFAR-100, WRN-37-2~\cite{zhou2020time} for STL-10, and ResNet-50~\cite{he2016deep} for ImageNet. As for training parameters, we follow the unified codebase TorchSSL~\cite{zhang2021flexmatch} to ensure fair comparisons. Specifically, batch sizes $B_L$ and $B_U$ are set to 128 and 128 on ImageNet and 64 and 448 on the remaining datasets. Moreover, the model is trained using the SGD optimizer with an initial learning rate of 0.03 and a momentum decay of 0.9. The learning rate is adjusted by a cosine decay scheduler over a total of $2^{20}$ steps. Besides, we train the EMA model with a momentum decay of 0.999 for inference. The fixed threshold $\tau$ in the first branch is set to 0.95. For SVHN, we follow previous studies~\cite{wang2022freematch, xu2021dash} and constrain the dynamic threshold within the range of [0.9, 1.0]. To suppress randomness, we repeat each experiment three times and report the average accuracy in \cref{tab:performance}. We omit the standard deviation given the stable performance of the proposed paradigm across different random seeds.

\textbf{Performance.} We apply the proposed paradigm~(IFMatch) to several algorithms and compare their performance with existing methods on multiple benchmarks. The results presented in \cref{tab:performance} underscore the significant contribution of our paradigm to promoting the performance of existing SSL algorithms. Specifically, the four algorithms involved achieve the average improvement of 4.05\%, 3.22\%, 1.62\%, and 1.39\% across all datasets. Moreover, previous methods struggle to enhance performance on less challenging tasks, such as CIFAR-10. In contrast, when integrated with the proposed paradigm, IFMatch~(Fix) outperforms FixMatch by 2.29\%/0.83\%/0.70\% on CIFAR-10 with 40/250/4000 labeled samples, respectively. Furthermore, our approach effectively addresses the challenges posed by extremely limited labeled samples, with IFMatch~(Fix) achieving enhancements of 14.51\% and 12.68\% on STL-10 with 40 labels and CIFAR-100 with 400 labels, respectively, compared to FixMatch. Moreover, the four algorithms achieve an average promotion of 3.69\%/2.30\% on ImageNet with 100k/400k labels when following the proposed paradigm, indicating its effectiveness on large-scale datasets. Additionally, when employing our paradigm, the performance gap between dynamic and constant thresholds considerably narrows on challenging tasks, while IFMatch~(Fix) even emerges as the preferred method on some less demanding tasks. The shifted performance gap implies a reduced reliance on complex threshold strategies of our paradigm, which primarily benefits from the expanded augmentation space. In summary, the impressive performance observed in \cref{tab:performance} confirms the effectiveness of the proposed paradigm.

\subsection{Imbalanced Semi-Supervised Learning}

\textbf{Settings.} We evaluate the proposed paradigm in the context of imbalanced SSL, where both labeled and unlabeled data exhibit a long-tailed distribution. Consistent with prior studies~\cite{lee2021abc, oh2022daso, wei2021crest, lai2022smoothed, fan2022cossl, chen2023softmatch}, we construct labeled and unlabeled sets using the configurations of $N_c=N_1\cdot\gamma^{-\frac{c-1}{C-1}}$ and $M_c=M_1\cdot\gamma^{-\frac{c-1}{C-1}}$. Specifically, for CIFAR-10-LT, we set $N_1$ to 1500, $M_1$ to 3000, and $\gamma$ to range from 50 to 150. Moreover, for CIFAR-100-LT, we set $N_1$ to 150, $M_1$ to 300, and $\gamma$ to span from 20 to 100. In all experiments, we employ WRN-28-2 as the backbone and utilize the Adam optimizer~\cite{kingma2014adam} with a weight decay of 4e-5. Besides, batch sizes $B_L$ and $B_U$ are fixed as 64 and 128 respectively. Additionally, the learning rate is initially set to 2e-3 and adjusted by a cosine decay scheduler during training. We set the fixed threshold $\tau$ in the first branch to 0.95. To account for randomness, we report the mean and standard deviation of the performance across three different random seeds.

\begin{table}[t]
    \scriptsize
    \centering
    \caption{Performance~(\%) on CIFAR-10/100-LT with varying imbalance ratio.}
    \begin{tabular}{c|ccc|ccc}
    \toprule
    Dataset & \multicolumn{3}{c}{CIFAR-10-LT} \vline & \multicolumn{3}{c}{CIFAR-100-LT} \\
    \midrule
    $\gamma$ & 50 & 100 & 150 & 20 & 50 & 100 \\
    \midrule
    FixMatch~\cite{sohn2020fixmatch} & 81.54{\tiny ±0.30} & 74.89{\tiny ±1.20} & 70.38{\tiny ±0.88} & 49.58{\tiny ±0.78} & 42.11{\tiny ±0.33} & 37.60{\tiny ±0.48} \\
    IFMatch~(Fix) & \underline{85.32{\tiny ±0.33}} & 79.46{\tiny ±0.35} & \underline{75.59{\tiny ±0.48}} & \underline{53.61{\tiny ±0.25}} & \underline{44.58{\tiny ±0.38}} & \textbf{39.92{\tiny ±0.37}} \\
    \midrule
    FlexMatch~\cite{zhang2021flexmatch} & 81.87{\tiny ±0.19} & 74.49{\tiny ±0.92} & 70.20{\tiny ±0.36} & 50.89{\tiny ±0.60} & 42.80{\tiny ±0.39} & 37.30{\tiny ±0.47} \\
    IFMatch~(Flex) & 84.36{\tiny ±0.12} & 79.04{\tiny ±0.42} & 73.02{\tiny ±0.47} & 51.83{\tiny ±0.33} & 43.49{\tiny ±0.35} & 37.98{\tiny ±0.36} \\
    \midrule
    SoftMatch~\cite{chen2023softmatch} & 83.45{\tiny ±0.29} & 77.07{\tiny ±0.37} & 72.60{\tiny ±0.46} & 51.91{\tiny ±0.55} & 43.76{\tiny ±0.51} & 38.92{\tiny ±0.81} \\
    IFMatch~(Soft) & \textbf{85.56{\tiny ±0.15}} & \textbf{80.28{\tiny ±0.24}} & \textbf{76.18{\tiny ±0.34}} & \textbf{53.97{\tiny ±0.32}} & \textbf{44.98{\tiny ±0.41}} & \underline{39.68{\tiny ±0.41}} \\
    \midrule
    FreeMatch~\cite{wang2022freematch} & 83.30{\tiny ±0.33} & 75.96{\tiny ±0.49} & 71.20{\tiny ±0.64} & 51.60{\tiny ±0.91} & 42.95{\tiny ±0.48} & 37.50{\tiny ±0.23} \\
    IFMatch~(Free) & 84.88{\tiny ±0.20} & \underline{80.12{\tiny ±0.29}} & 74.85{\tiny ±0.47} & 52.77{\tiny ±0.46} & 43.68{\tiny ±0.45} & 38.00{\tiny ±0.55}\\
    \bottomrule
    \end{tabular}
    \vspace{-0.2cm}
    \label{tab:imbalance}
\end{table}

\textbf{Performance.} We apply the proposed paradigm to numerous algorithms and compare the performance with their counterparts following the old paradigm. The results in \cref{tab:imbalance} demonstrate that our paradigm substantially improves existing models, enabling them to achieve state-of-the-art performance. Specifically, when integrated with the enhanced paradigm, the four algorithms achieve the average promotion of 3.72\%, 2.03\%, 2.16\%, and 1.92\% across all benchmarks. Notably, IFMatch~(Fix) outperforms FixMatch by 5.21\% and 2.32\% on CIFAR-10-LT~($\gamma$=150) and CIFAR-100-LT~($\gamma$=100), highlighting the robustness of the proposed paradigm when facing severe imbalance. Moreover, the performance gap between dynamic-threshold-based methods and FixMatch substantially narrows within our paradigm. Accordingly, the proposed approach promotes the utilization of unlabeled data, reducing reliance on complex threshold mechanisms. Overall, the impressive performance observed in imbalanced SSL suggests the potential of our approach to be deployed in real-world applications.

\subsection{Component Analysis}

This section presents the contribution of each component within the proposed paradigm. Unless otherwise stated, we conduct experiments on IFMatch~(Fix). Additionally, \textit{we provide running speed analysis, feature visualization results~\cite{van2008visualizing}, and performance on semi-supervised semantic segmentation in the Appendix}. The subsequent analysis presents the revealed findings.

\begin{table}[t]
    \scriptsize
    \centering
    \tabcolsep=0.1cm
    \caption{Experiments~(\%) on the strategies for integrating image-level and feature-level perturbations. CBI abbreviates the confidence-based identification strategy. CIFAR-$m$-$n$ denotes CIFAR-$m$ with $n$ labeled samples. Line 3 combines the structure setting of UniMatch~\cite{yang2023revisiting} and our proposed feature-level perturbation techniques.}
    \begin{tabular}{c|c|c|c}
    \toprule
    Branch I & Branch II & CIFAR-10-40 & CIFAR-100-400 \\
    \midrule
    \multicolumn{2}{c}{FixMatch~(baseline, no $\mathcal{A}^\mathcal{F}$)} \vline & 92.53 & 53.58 \\
    \multicolumn{2}{c}{$\mathcal{A}^{\mathcal{F}_s}$ + $\mathcal{A}^{\mathcal{I}_s}$ in one branch} \vline & 88.26 & 51.47 \\
    \midrule
    $\mathcal{A}^{\mathcal{F}_s}$ & $\mathcal{A}^{\mathcal{I}_s}$ & 95.53 & 64.54 \\
    $\mathcal{A}^{\mathcal{F}_s}$ + $\mathcal{A}^{\mathcal{I}_w}$ & $\mathcal{A}^{\mathcal{F}_w}$ + $\mathcal{A}^{\mathcal{I}_s}$ & 95.47 & 63.56\\
    $\mathcal{A}^{\mathcal{F}_s}$ + $\mathcal{A}^{\mathcal{I}_w}$ & $\mathcal{A}^{\mathcal{F}_w}$ + $\mathcal{A}^{\mathcal{I}_s}$ + CBI & 95.82 & 66.26\\
    \bottomrule
    \end{tabular}
    \label{tab:ablation_structure}
\end{table}

\textbf{The proposed approach to integrating image-level and feature-level perturbations yields optimal performance.} As the simplest idea, we can integrate $\mathcal{A}^{\mathcal{F}_s}$ and $\mathcal{A}^{\mathcal{I}_s}$ within one branch to establish a dual-branch structure similar to the old paradigm. Nonetheless, such a toy design engenders disruptive perturbations that exceed acceptable augmentation levels for consistency regulation, resulting in the performance degradation presented in lines 1-2 of \cref{tab:ablation_structure}. An alternative approach, proposed by UniMatch~\cite{yang2023revisiting}, combines $\mathcal{A}^\mathcal{F}$ and $\mathcal{A}^\mathcal{I}$ in separate branches, each employing a distinct type of perturbation (line 3). Despite its effectiveness, this method fails to boost interactions between the two types of perturbations. Accordingly, we attempt to achieve the co-occurrence of $\mathcal{A}^\mathcal{F}$ and $\mathcal{A}^\mathcal{I}$ within one branch, thereby constructing a more comprehensive augmentation space. The proposed paradigm adopts a milder combination strategy, specifically using $\mathcal{A}^{\mathcal{F}_s} + \mathcal{A}^{\mathcal{I}_w}$ and $\mathcal{A}^{\mathcal{F}_w} + \mathcal{A}^{\mathcal{I}_s}$ in the two student branches respectively. However, as shown in line 4 of \cref{tab:ablation_structure}, the introduction of $\mathcal{A}^{\mathcal{F}_w}$ leads to a performance drop compared to UniMatch, probably due to the derived difficulties in recognizing hard samples. To address this issue, we devise a confidence-based identification strategy to distinguish between naive and challenging samples, posing additional challenges exclusively for naive samples. Finally, the proposed paradigm~(IFMatch), as presented in line 5, achieves the optimal performance among all candidate designs for integrating $\mathcal{A}^\mathcal{F}$ and $\mathcal{A}^\mathcal{I}$, demonstrating its effectiveness and advantages.

\begin{table}[t]
    \scriptsize
    \centering
    \tabcolsep=0.1cm
    \caption{Ablation study~(\%) for the strategies of feature-level perturbation.}
    \begin{tabular}{c|c|c}
    \toprule
    $\mathcal{A}^\mathcal{F}$ Strategy & CIFAR-10-40 & CIFAR-100-400 \\ 
    \midrule
    All strategies & \textbf{95.82} & \textbf{66.26} \\
    \textit{w/o} channel-wise dropout & 95.66 ($\downarrow$0.16) & 65.68 ($\downarrow$0.58) \\
    \textit{w/o} spatial-wise dropout & 95.57 ($\downarrow$0.25) & 65.77 ($\downarrow$0.49) \\
    \textit{w/o} translation & 95.59 ($\downarrow$0.23) & 65.70 ($\downarrow$0.56) \\
    \textit{w/o} shearing & 95.58 ($\downarrow$0.24) & 65.81 ($\downarrow$0.47) \\
    \textit{w/o} value modification & 95.62 ($\downarrow$0.20) & 65.75 ($\downarrow$0.51) \\
    \bottomrule
    \end{tabular}
    \label{tab:ablation_strategy}
\end{table}

\textbf{All feature-level perturbation strategies contribute to the overall performance.} As shown in \cref{tab:ablation_strategy}, we conduct an ablation study on feature-level perturbation strategies. The baseline model in line~1 integrates all proposed strategies and achieves state-of-the-art performance. In subsequent configurations~(line~2-6), we sequentially remove each strategy. Compared to the baseline, the absence of any proposed perturbation technique leads to a decline in performance on both datasets. Therefore, the proposed perturbation strategies exhibit a synergistic effect, collaboratively exploring the feature perturbation space.

\begin{table}[t]
    \scriptsize
    \centering
    \tabcolsep=0.1cm
    \caption{Exploration~(\%) for the threshold strategy in the first branch.}
    \begin{tabular}{c|c|c|c}
    \toprule
    Branch I & Branch II & CIFAR-10-40 & CIFAR-100-400 \\
    \midrule
    FlexMatch & FlexMatch & 95.28 & 63.38 \\
    $\tau=0.95$ & FlexMatch & \textbf{95.90} & \textbf{66.28} \\
    \midrule
    SoftMatch & SoftMatch & 95.39 & 63.80 \\
    $\tau=0.95$ & SoftMatch & \textbf{95.84} & \textbf{67.02} \\
    \midrule
    FreeMatch & FreeMatch & 95.56 & 64.45 \\
    $\tau=0.95$ & FreeMatch & \textbf{95.83} & \textbf{66.65} \\
    \bottomrule
    \end{tabular}
    \label{tab:threshold_branchI}
\end{table}

\textbf{Feature-level perturbation $\mathcal{A}^\mathcal{F}$ favors a high constant threshold over a dynamic one.} As detailed in \cref{sec: new paradigm}, $\mathcal{A}^\mathcal{F}$ and $\mathcal{A}^\mathcal{I}$ may require different threshold mechanisms due to their distinct characteristics. Considering the central role of $\mathcal{A}^\mathcal{F}$~($\mathcal{A}^{\mathcal{F}_s}$) in the first student branch, we further delve into the threshold strategy best suited for it. In general, existing threshold mechanisms can be categorized into constant thresholds and dynamic thresholds. The former prioritizes high pseudo-label accuracy at the expense of diminished utilization of unlabeled data, while the latter performs conversely. Accordingly, we determine the optimal threshold for the first branch from two candidates: a dynamic threshold mirroring the one used in the second branch and the constant threshold used in FixMatch. Combining \cref{tab:threshold_branchI} and \cref{tab:performance}, the proposed paradigm consistently outperforms the old paradigm regardless of the threshold adopted in the first branch. Additionally, the constant threshold beats all dynamic threshold mechanisms, suggesting that feature-level perturbation $\mathcal{A}^\mathcal{F}$ is better suited to a constant threshold of 0.95, rather than existing dynamic threshold mechanisms.

\begin{table}[t]
    \scriptsize
    \centering
    \tabcolsep=0.1cm
    \caption{Comparison~(\%) between SAA~\cite{gui2023enhancing} and confidence-based identification.}
    \begin{tabular}{c|c|c}
    \toprule
    Identification Strategy & CIFAR-10-40 & CIFAR-100-400 \\
    \midrule
    SAA~\cite{gui2023enhancing} & 95.54 & 65.32 \\
    Confidence-Based Identification & \textbf{95.82} & \textbf{66.26} \\
    \bottomrule
    \end{tabular}
    \label{tab:ablation_combination}
\end{table}

\begin{figure}[t]
  \centering
  \includegraphics[width=0.75\textwidth]{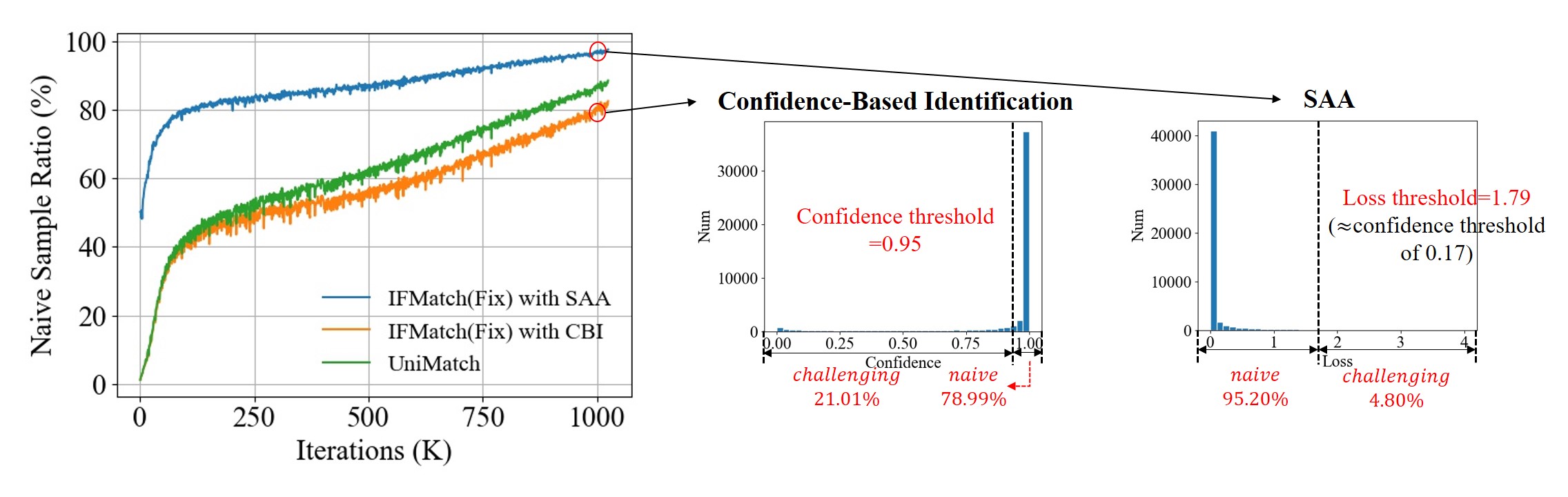}
  \vspace{-0.1cm}
  \caption{Visualization for naive sample ratio and the identification process of SAA~\cite{gui2023enhancing} and confidence-based identification on CIFAR-100-400.}
  \label{fig:adaptive}
\end{figure}

\textbf{IFMatch outperforms previous methods from the perspective of the naive sample ratio.} Firstly, as shown in the left part of \cref{fig:adaptive}, our approach~(orange) achieves a lower naive sample ratio than UniMatch~(green). This suggests more effective utilization of unlabeled data in our paradigm, primarily benefiting from combining both types of perturbation within each branch. Secondly, we compare the confidence-based identification strategy and SAA~\cite{gui2023enhancing} within our paradigm to discern a better naive sample identification method. Specifically, SAA utilizes loss values as the criteria, recording historical loss for each sample and performing naive-challenging division by OTSU~\cite{otsu1979threshold}. As shown in \cref{tab:ablation_combination}, the proposed module consistently outperforms SAA. Additionally, the naive sample ratio presented in the left part of \cref{fig:adaptive} indicates that SAA~(blue) tends to produce identification biased toward naive samples~(\eg starting with a ratio of $0.5$ in the first iteration). Furthermore, we analyze the identification process of both strategies in the later training stages, as shown in the right part of \cref{fig:adaptive}. Notably, when employing the cross-entropy loss, the loss threshold of 1.79 in SAA corresponds to a confidence threshold of 0.17~(computed as $e^{-1.79}$), which is a detrimental configuration. The limitations of SAA can be ascribed to the ill-suited nature of the OTSU technique for segmenting a histogram with a monotonically decreasing trend. In contrast, our strategy remains effective in distinguishing between naive and challenging samples even if the unlabeled set is predominantly composed of naive samples.

\begin{figure}[t]
    \centering
    \includegraphics[width=0.75\textwidth]{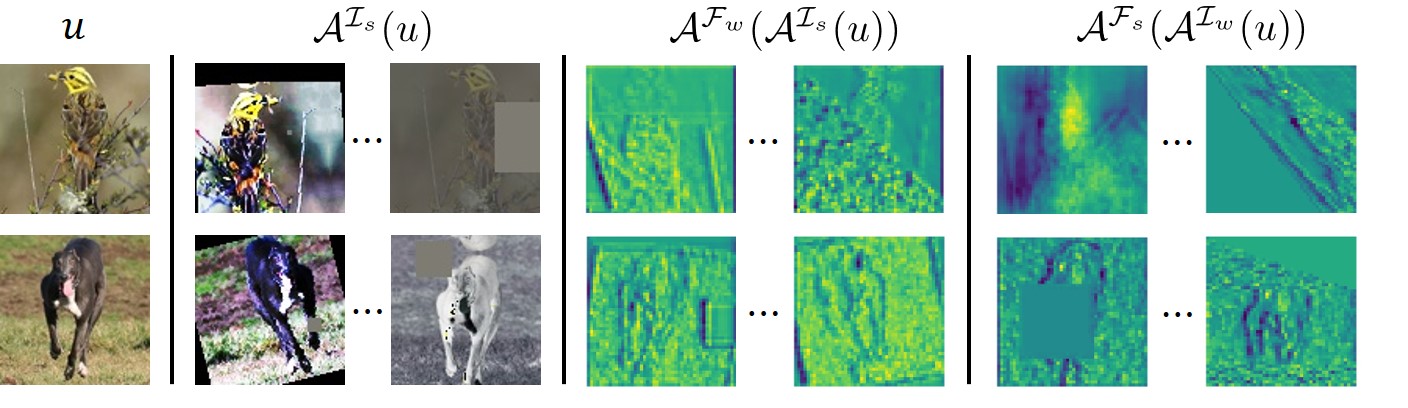}
    \vspace{-0.1cm}
    \caption{Visualization of perturbations. We provide the feature map for samples that undergo feature-level augmentation. A naive sample $u$ exhibits the characteristic of $\forall \mathcal{A}^{\mathcal{I}_s}(u), \mathcal{H}(\hat{p}^{\mathcal{I}_w}_i, p^{\mathcal{I}_s}_i) \approx 0$, thus contributing trivially to the model's performance.}
    \label{fig:vis}
\end{figure}

\textbf{Visualization analysis.} We present the visualization for image-level and feature-level perturbations in \cref{fig:vis}, where the four parts~(from left to right) correspond to the raw images, the images subjected to $\mathcal{A}^{\mathcal{I}_s}$, and the feature maps that undergo $\mathcal{A}^{\mathcal{F}_w} \circ \mathcal{A}^{\mathcal{I}_s}$ and $\mathcal{A}^{\mathcal{F}_s} \circ \mathcal{A}^{\mathcal{I}_w}$. We can observe that the exclusive reliance on $\mathcal{A}^{\mathcal{I}_s}$ engenders easy-to-recognize~(naive) samples. In contrast, the proposed paradigm effectively combines $\mathcal{A}^\mathcal{I}$ and $\mathcal{A}^\mathcal{F}$ to expand the augmentation space. \textit{For additional visualization results, please refer to Appendix Sec.2.}
\section{Conclusion}
\label{sec:conclusion}
This paper introduces feature-level perturbation to expand the augmentation space of the conventional semi-supervised learning framework, establishing the image-feature weak-to-strong consistency paradigm~(IFMatch). To achieve effective feature-level perturbation, we propose refined designs that consider both perturbation position and strategy, facilitating diversity in terms of intensity and form. Furthermore, our paradigm develops two parallel student branches to seamlessly integrate the two types of perturbations and comprehensively explore the augmentation space. Additionally, we devise a confidence-based identification strategy to distinguish between naive and challenging samples, posing additional challenges exclusively for naive samples. Extensive experiments are conducted on multiple benchmarks, including both balanced and imbalanced label distributions. The experimental results demonstrate the effectiveness of our approach.

%
%
\bibliographystyle{splncs04}
\bibliography{main}

\clearpage
\appendix

\section{Feature-Level Perturbation Strategies}

As mentioned in Sec.3.2 of the main paper, we develop a series of feature-level perturbation strategies from different perspectives, including `dropout'~(comprising channel-wise dropout and spatial-wise dropout), `movement'~(encompassing translation and shearing in both the X-axis and Y-axis), and `value'~(involving the weighted sum of the smoothed and input feature maps). In this section, we will elaborate on the implementation of these strategies. For the sake of simplicity, we maintain a consistent notation for all perturbation strategies.
\begin{align}
    f^{out} &= \Phi(f^{in}) \\
    f^{out}, f^{in} &\in R^{C \times H \times W}
\end{align}
where $f^{in}$ and $f^{out}$ denote the input and output feature maps, $\Phi$ stands for the perturbation operation, $C$ represents the number of channels, $H$ and $W$ signifies the height and width of the feature maps. 

\subsection{Channel-Wise Dropout}
This perturbation strategy applies channel-wise dropout with a drop probability of $0.5$ to hidden representations. Mathematically, the feature-level perturbation using channel-wise dropout can be defined as follows.
\begin{align}
    f^{out} &= \text{torch.nn.Dropout2d}(p=0.5)(f^{in})
\end{align}

\subsection{Spatial-Wise Dropout}
Spatial-wise dropout discards a randomly selected rectangle region across all channels, similar to the Cutout operation applied to raw images. To determine the dropped spatial region, the strategy performs the following steps. Firstly, we compute the height $h_d$ and width $w_d$ of the dropped region by rescaling the size of input feature maps. Subsequently, we randomly select the upper left point $(x, y)$ of the dropped area. With the size and the location of its top-left point in place, we can uniquely define the dropped area. Mathematically speaking, the above process can be expressed as follows.
\begin{align}
    h, w &= \text{int}(\alpha \times H), \text{int}(\alpha \times W) \\
    x, y &= \mathcal{U}[0, H - h], \mathcal{U}[0, W - w]
\end{align}
In all experiments, we set $\alpha$ to $0.5$. Once the dropped spatial region is determined, we generate a 0-1 mask denoted as $m \in R^{H \times W}$ and implement spatial-wise dropout by multiplying the mask with the input feature maps. Formally, the output feature maps can be computed as follows.
\begin{align}
    m(i, j) &= 
    \begin{cases}
        0 &\textit{if }(i, j) \textit{ in Rectangle}(x, y, h, w) \\
        1 &\textit{otherwise}
    \end{cases}
    \\
    f^{out} &= f^{in} \odot m
\end{align}
where $\odot$ denotes the element-wise product. Note that a similar rescaling approach is applied to the remaining feature values as in channel-wise dropout. 

\begin{figure}[t]
    \centering
    \includegraphics[width=0.6\textwidth]{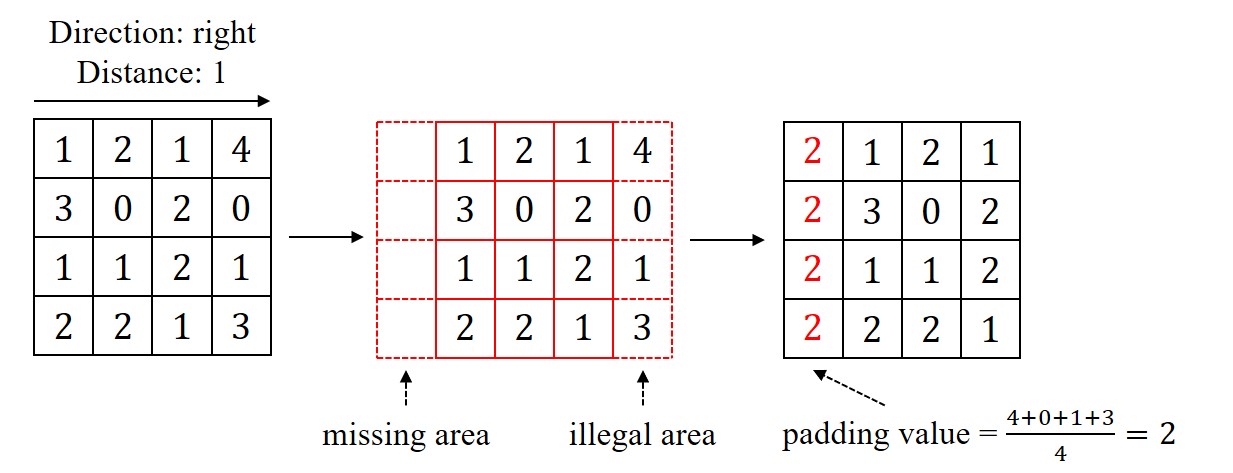}
    \caption{A toy sample for feature-level translation perturbation.}
    \label{fig:translation}
\end{figure}

\subsection{Translation}
The feature-level translation perturbation randomly selects a direction from the candidate set~(up, down, left, and right) with equal probability and translates the input feature map along the determined direction. The length of the translation path denoted as $l$ can be computed by rescaling the size of the input feature maps. Specifically, we first draw a random factor $\alpha$ from a uniform distribution ranging from $0$ to $\alpha_{max}$ and then compute the translation distance. Taking the translation along the X-axis as an example, the above process can be mathematically described as follows.
\begin{align}
    \alpha &\sim \mathcal{U}[0, \alpha_{max}] \\
    l &= \text{int}(\alpha \times W)
\end{align}
Throughout all experiments, we set $\alpha_{max}$ to $0.5$. Given the translation direction and distance, the strategy assigns the corresponding value from the input feature maps to the output feature maps. Furthermore, to address the areas left vacant due to the translation operation, we utilize the average value outside the legal region as padding, thereby avoiding significant information loss and ensuring numerical stability. For a more intuitive understanding of the translation operation, please refer to the toy example presented in \cref{fig:translation}.

\begin{figure}[t]
    \centering
    \includegraphics[width=0.6\textwidth]{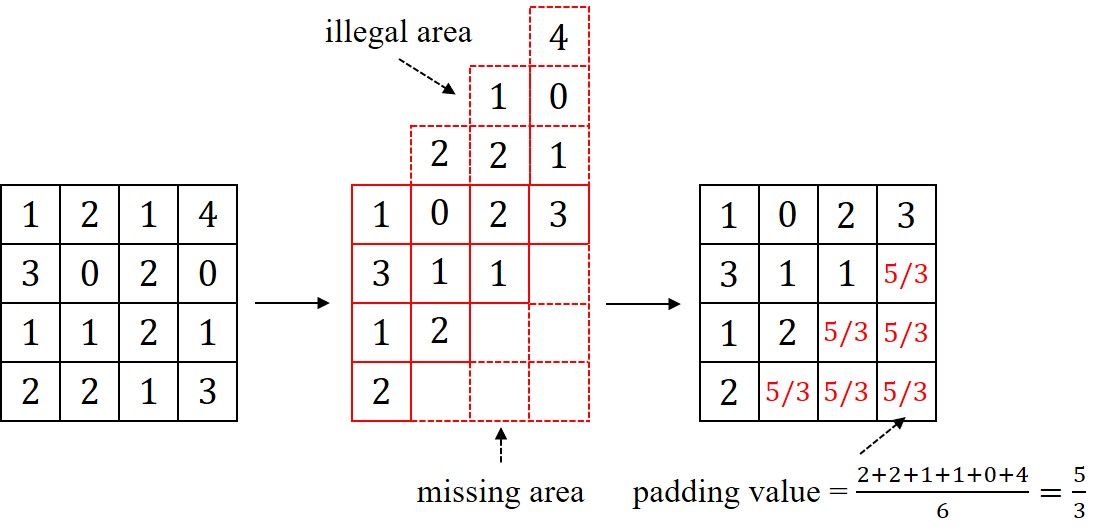}
    \caption{A toy sample for feature-level shearing perturbation.}
    \label{fig:shearing}
\end{figure}

\subsection{Shearing}
Similar to translation, the shearing operation randomly selects one direction from a set of candidate directions, each with equal probability. The feature map is then sheared along the chosen direction. The length of the shearing path denoted as $l$ can be calculated by rescaling the size of the input feature maps, and intermediate distances can be obtained by linear interpolation. Specifically, we first draw a random factor $\alpha$ from a uniform distribution between $0$ and $\alpha_{max}$ and then compute the shearing distance. Taking the shearing operation in the X-axis as an illustrative example, the above process can be expressed as follows.
\begin{align}
    \alpha &\sim \mathcal{U}[0, \alpha_{max}] \\
    l &= \text{int}(\alpha \times W) \\
    [l_1, \cdots, l_W] &= \text{torch.linaspace}(0, l, W)
\end{align}
In all experiments, we set $\alpha_{max}$ to $1.0$. Subsequently, the strategy assigns corresponding values from the input feature maps to the output feature maps and employs the average value of the illegal region to pad the missing areas. \cref{fig:shearing} provides an example of the shearing operation.

\subsection{Value Modification}
The value modification operation aims to change the feature value while preserving the relationship within each feature map. To achieve this goal, we first employ a random-sized group convolution to facilitate local smoothing in each feature map. The random-sized configuration allows for various degrees of smoothness, thus comprehensively exploring the feature perturbation space. The size of the kernel $k$ is randomly sampled from the set of all odd numbers within the range of $3$ to min$(H, W)$ with equal probability. Subsequently, we adopt the average smoothing convolution to encode the input feature maps. Finally, we draw a random factor $\alpha$ from a uniform distribution between $\alpha_{min}$ and $\alpha_{max}$ and use the weighted sum between the input and smoothed feature maps as the perturbation results. Mathematically speaking, the above process can be expressed as follows.
\begin{align}
    k &\sim \{\mathcal{U}[3, \text{min}(H, W)], \text{odd number}\} \\
    K[i, j] &= \frac{1}{k^2} \\
    \alpha &\sim \mathcal{U}[\alpha_{min}, \alpha_{max}] \\
    f^{out} &= \alpha \times (K * f^{in}) + (1 - \alpha) \times f^{in}
\end{align}
where $*$ denotes the convolution operation. In all experiments, we set $\alpha_{min}$ and $\alpha_{max}$ to 0.50 and 0.95, respectively.

\subsection{Conclusion for Strategies}
The aforementioned perturbation strategies consider three distinct perspectives (`dropout', `movement', and `value'), thereby achieving diverse perturbation forms and providing a collective exploration of the feature perturbation space. Moreover, the parameter settings for each perturbation strategy are relatively simple. For example, the first four perturbation strategies all involve one single hyperparameter, and we set it to $0.5$ on the first three, indicating satisfactory consistency. Note that the configuration setting $\alpha_{max}$ to $1.0$ in the shearing operation leads to the same upper bound~(half the size of the feature map) for the missing area as the translation operation. Additionally, the value modification operation introduces two hyperparameters, $\alpha_{min}$ and $\alpha_{max}$, and our settings keep consistent with the counterparts in strong image-level perturbation.

\begin{figure*}[t]
  \centering
  \begin{subfigure}{0.24\textwidth}
    \includegraphics[width=\textwidth]{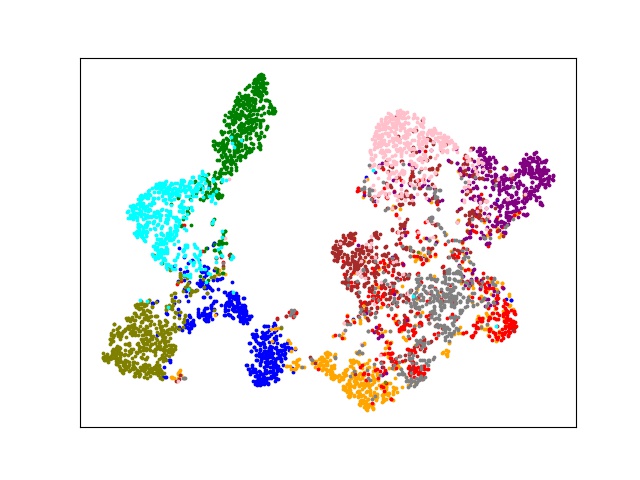}
    \caption{FixMatch on unlabeled set.}
    \label{fig:fix_unlabeled_set}
  \end{subfigure}
  \begin{subfigure}{0.24\textwidth}
    \includegraphics[width=\textwidth]{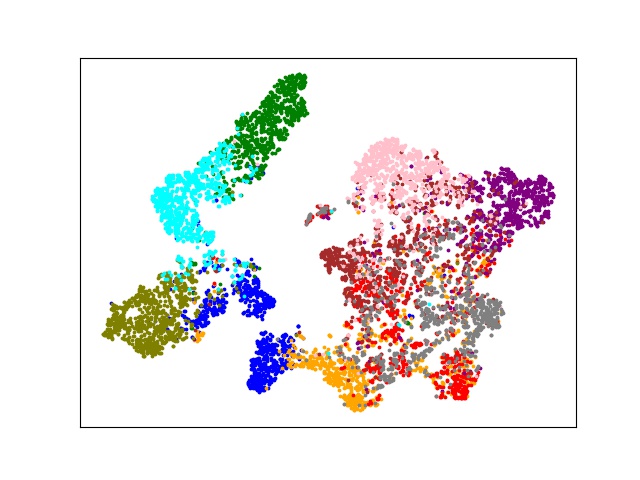}
    \caption{FixMatch on test set.}
   \label{fig:fix_test_set}
  \end{subfigure}
  \begin{subfigure}{0.24\textwidth}
    \includegraphics[width=\textwidth]{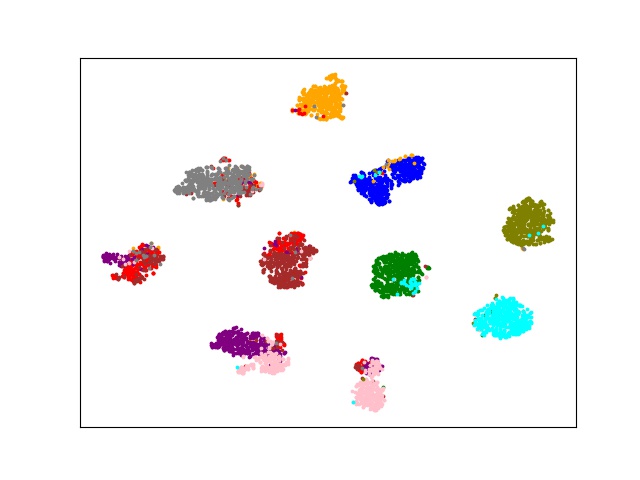}
    \caption{IFMatch~(Fix) on unlabeled set.}
    \label{fig:iffix_unlabeled_set}
  \end{subfigure}
  \begin{subfigure}{0.24\textwidth}
    \includegraphics[width=\textwidth]{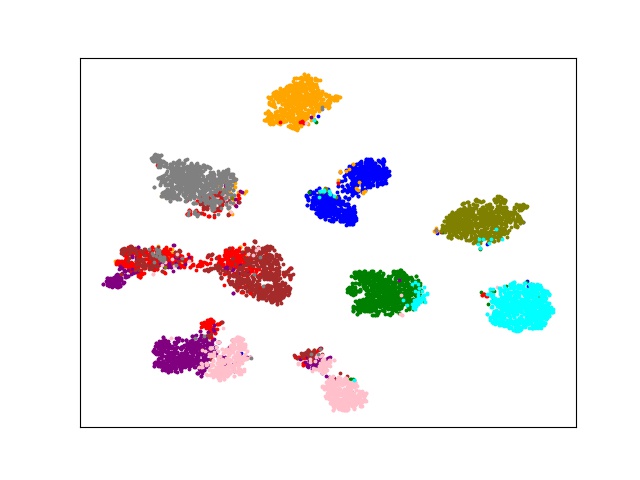}
    \caption{IFMatch~(Fix) on test set.}
   \label{fig:iffix_test_set}
  \end{subfigure}
  
  \begin{subfigure}{0.24\textwidth}
    \includegraphics[width=\textwidth]{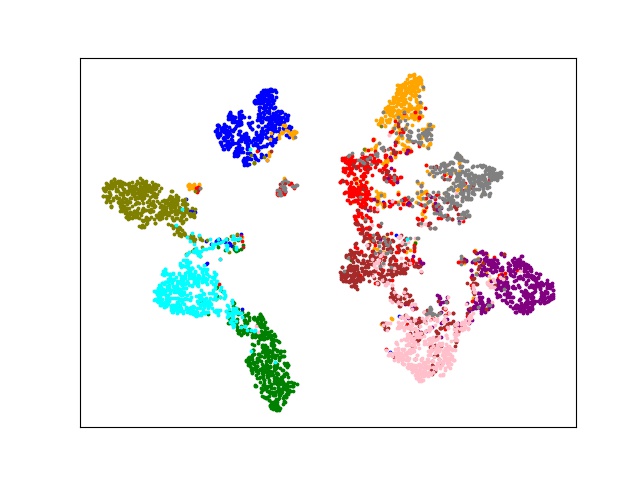}
    \caption{FlexMatch on unlabeled set.}
   \label{fig:flex_unlabeled_set}
  \end{subfigure}
  \begin{subfigure}{0.24\textwidth}
    \includegraphics[width=\textwidth]{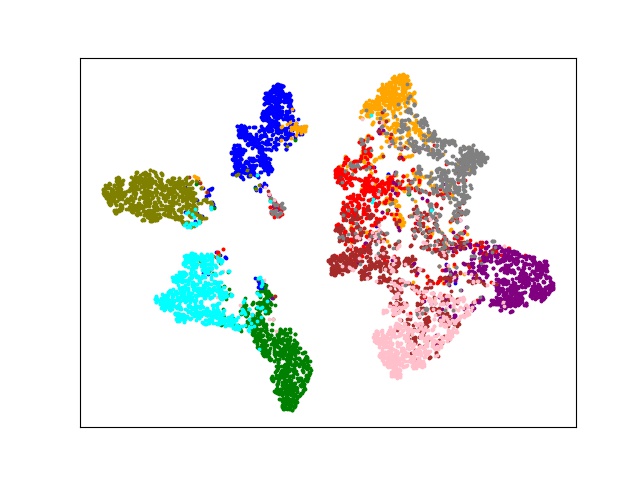}
    \caption{FlexMatch on test set.}
   \label{fig:flex_test_set}
  \end{subfigure}
  \begin{subfigure}{0.24\textwidth}
    \includegraphics[width=\textwidth]{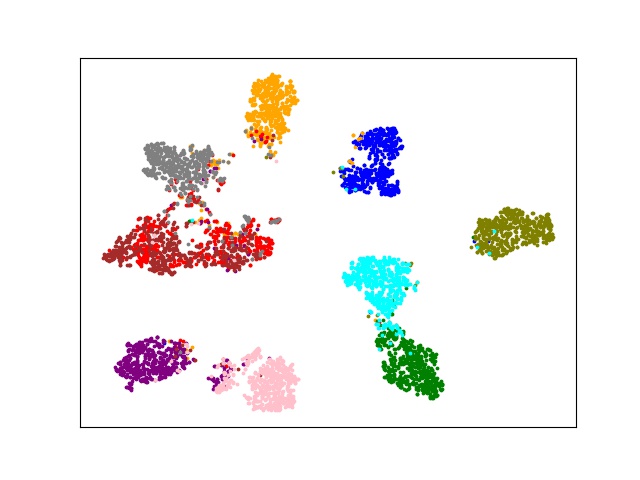}
    \caption{IFMatch~(Flex) on unlabeled set.}
    \label{fig:ifflex_unlabeled_set}
  \end{subfigure}
  \begin{subfigure}{0.24\textwidth}
    \includegraphics[width=\textwidth]{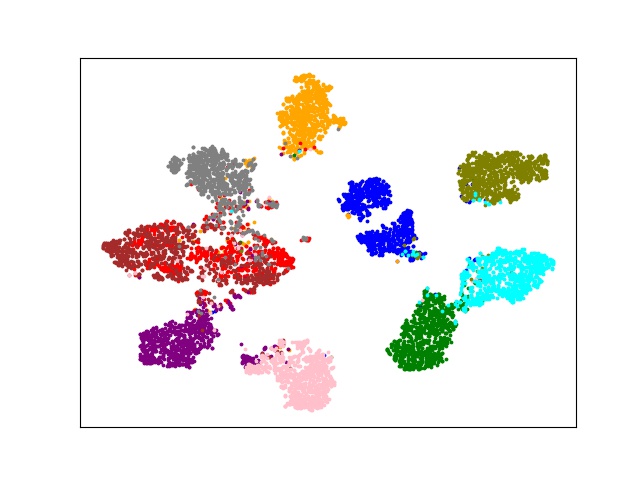}
    \caption{IFMatch~(Flex) on test set.}
   \label{fig:ifflex_test_set}
  \end{subfigure}

  \begin{subfigure}{0.24\textwidth}
    \includegraphics[width=\textwidth]{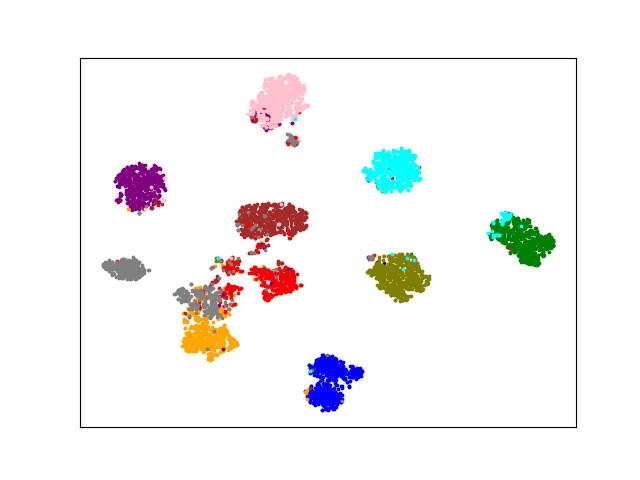}
    \caption{SoftMatch on unlabeled set.}
   \label{fig:soft_unlabeled_set}
  \end{subfigure}
  \begin{subfigure}{0.24\textwidth}
    \includegraphics[width=\textwidth]{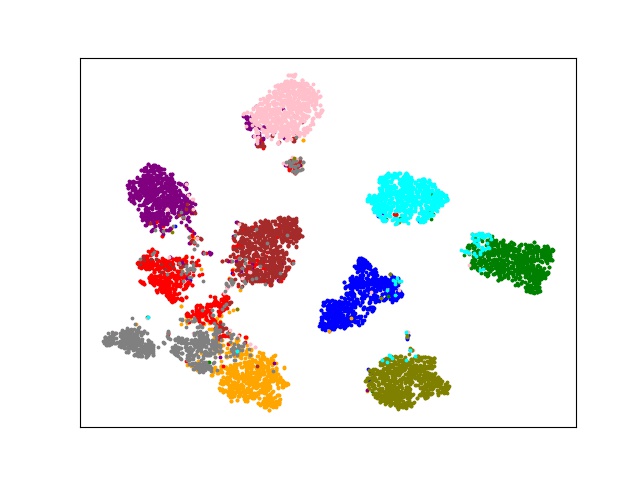}
    \caption{SoftMatch on test set.}
   \label{fig:soft_test_set}
  \end{subfigure}
  \begin{subfigure}{0.24\textwidth}
    \includegraphics[width=\textwidth]{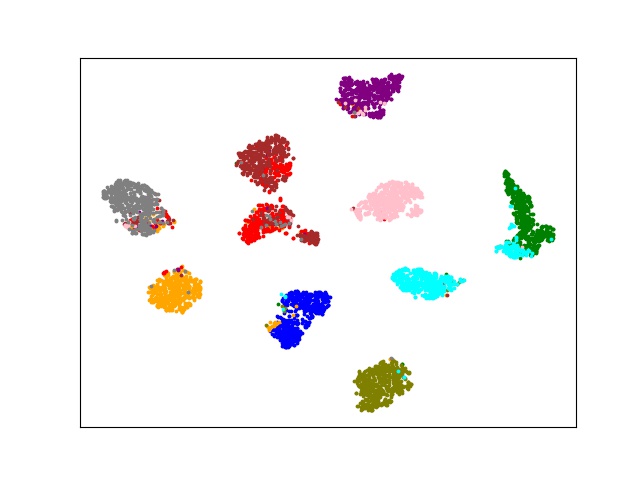}
    \caption{IFMatch~(Soft) on unlabeled set.}
    \label{fig:ifsoft_unlabeled_set}
  \end{subfigure}
  \begin{subfigure}{0.24\textwidth}
    \includegraphics[width=\textwidth]{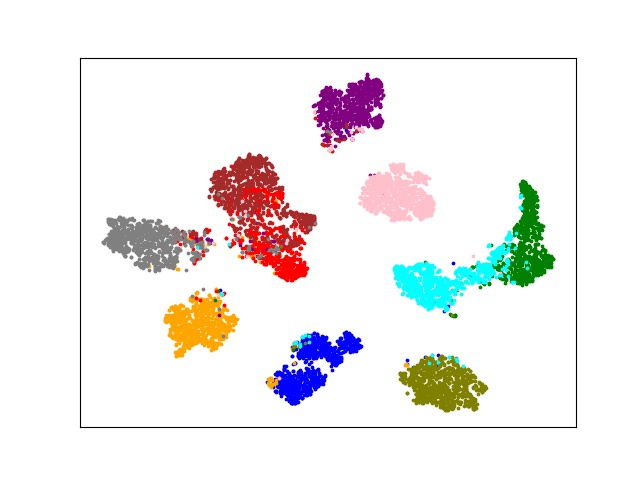}
    \caption{IFMatch~(Soft) on test set.}
   \label{fig:ifsoft_test_set}
  \end{subfigure}

  \begin{subfigure}{0.24\textwidth}
    \includegraphics[width=\textwidth]{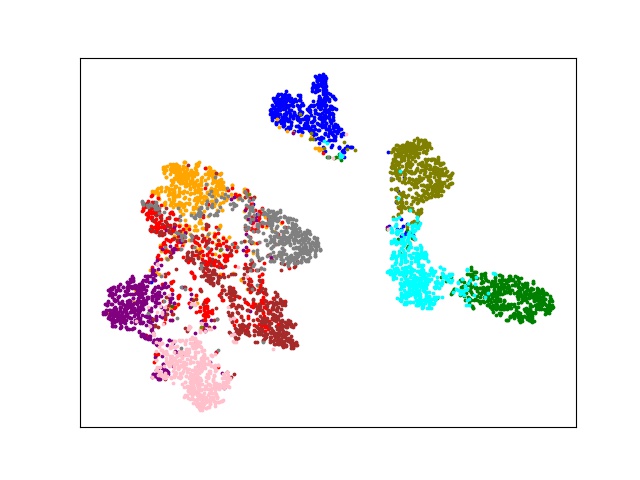}
    \caption{FreeMatch on unlabeled set.}
   \label{fig:free_unlabeled_set}
  \end{subfigure}
  \begin{subfigure}{0.24\textwidth}
    \includegraphics[width=\textwidth]{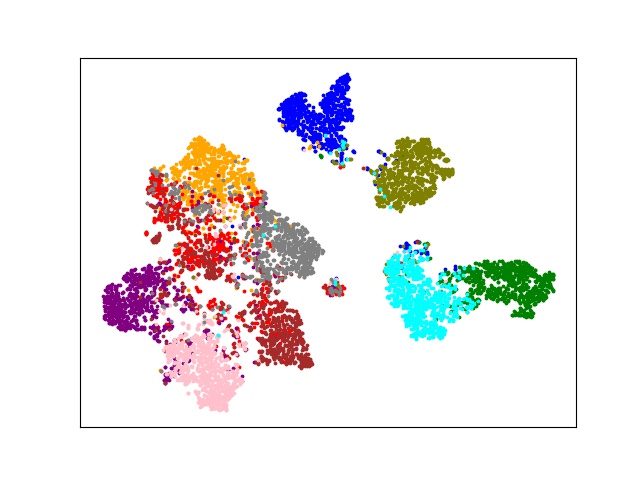}
    \caption{FreeMatch on test set.}
   \label{fig:free_test_set}
  \end{subfigure}
  \begin{subfigure}{0.24\textwidth}
    \includegraphics[width=\textwidth]{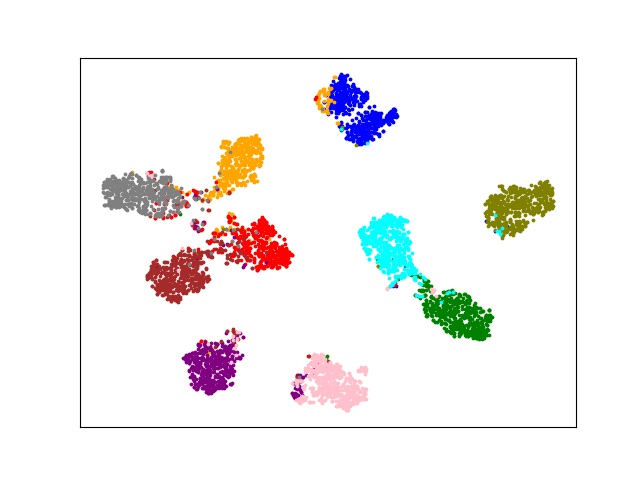}
    \caption{IFMatch~(Free) on unlabeled set.}
    \label{fig:iffree_unlabeled_set}
  \end{subfigure}
  \begin{subfigure}{0.24\textwidth}
    \includegraphics[width=\textwidth]{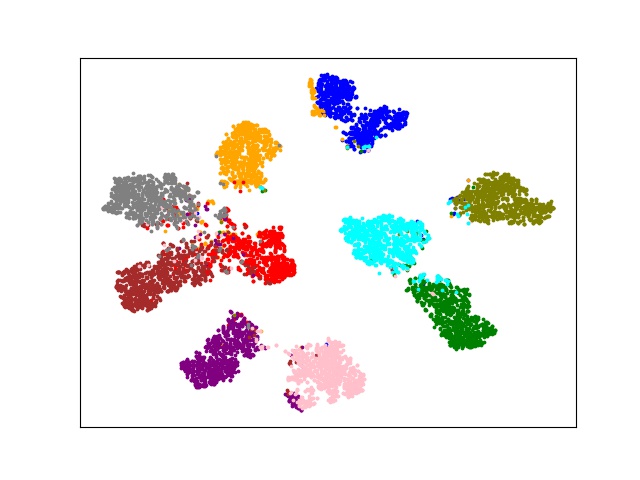}
    \caption{IFMatch~(Free) on test set.}
   \label{fig:iffree_test_set}
  \end{subfigure}
  \caption{Feature visualization of four algorithms when following the old and proposed paradigms on STL-10 with 40 labeled samples.}
  \label{fig:tsne}
\end{figure*}

\section{Visualization Analysis}

\subsection{High-Dimensional Features}
To diagnose the proposed paradigm in a more intuitive manner, we visualize the high-dimensional features on STL-10 with 40 labeled samples using T-SNE. The results are presented in \cref{fig:tsne}, wherein the first two and last two columns respectively depict the performance of algorithms following the old and proposed paradigms. To elaborate, algorithms adhering to the conventional approach typically generate loose and adjacent feature clusters. In contrast, the proposed paradigm achieves more separable and tightly clustered features, indicating the effectiveness of feature-level perturbation. Moreover, traditional algorithms often overfit noisy pseudo-labels, while the proposed approach positions most ambiguous samples near the decision boundary, minimizing their impacts on the model. Overall, the proposed image-level weak-to-strong consistency paradigm~(IFMatch) generates easy-to-distinguish features, laying a solid foundation for a robust classifier.

\begin{figure}[t]
    \centering
    \includegraphics[width=0.75\textwidth]{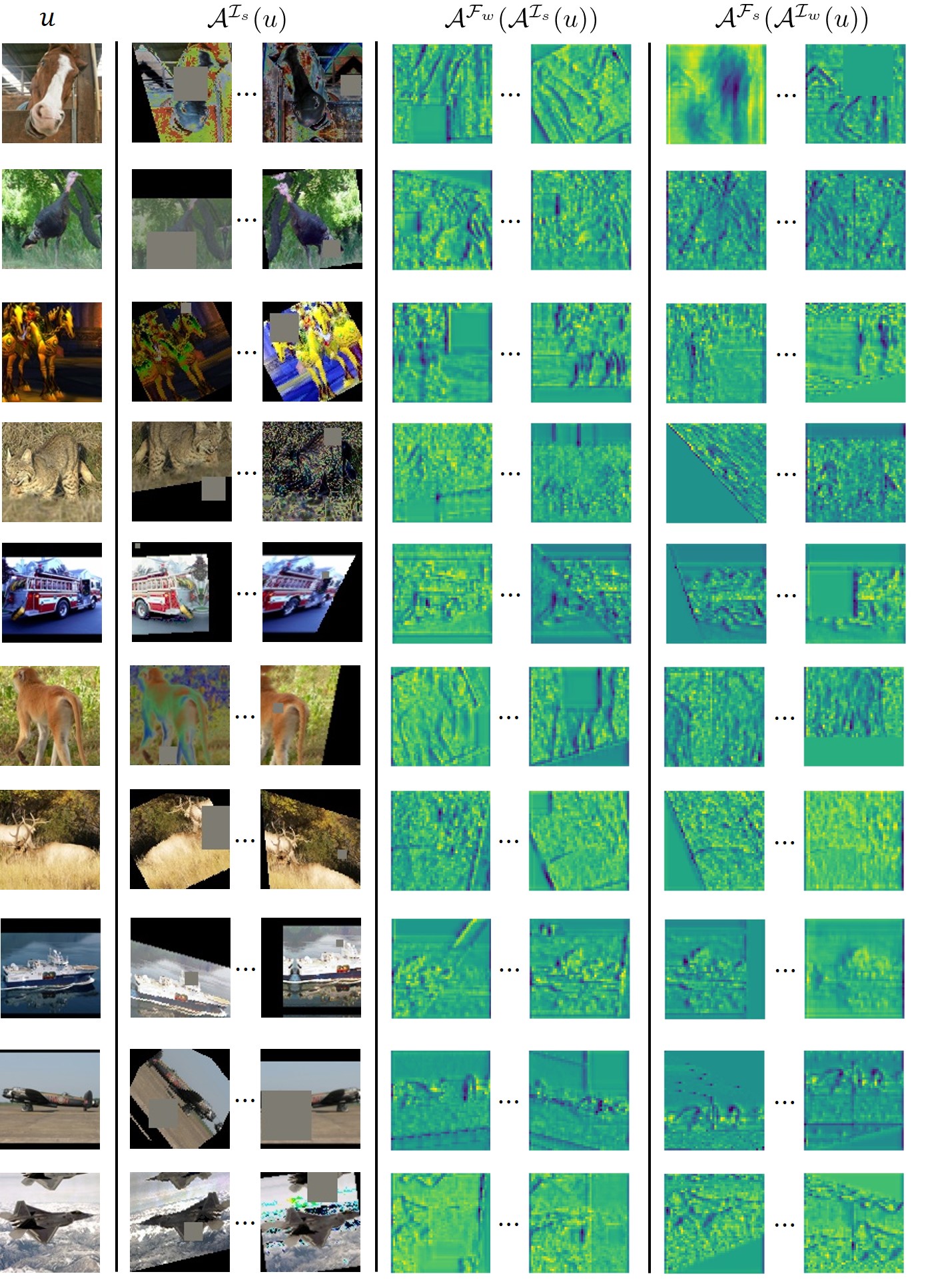}
    \caption{Visualization for image-level and feature-level perturbations. We provide the feature maps for samples that undergo feature-level perturbation.}
    \label{fig:vis_sup}
\end{figure}

\subsection{Image-Level Perturbation and Feature-Level Perturbation}
We present the visualization for image-level and feature-level perturbations in \cref{fig:vis_sup}, which serves as the supplement for the counterpart in the main paper. The four parts~(from left to right) in \cref{fig:vis_sup} correspond to the raw images, the images subjected to $\mathcal{A}^{\mathcal{I}_s}$, and the feature maps that undergo $\mathcal{A}^{\mathcal{F}_w} \circ \mathcal{A}^{\mathcal{I}_s}$ and $\mathcal{A}^{\mathcal{F}_s} \circ \mathcal{A}^{\mathcal{I}_w}$, respectively. As we can observe, a significant proportion of samples in the second part can be effortlessly classified, indicating the naive sample issue caused by the exclusive reliance on $\mathcal{A}^{\mathcal{I}_s}$. In contrast, the combination of $\mathcal{A}^\mathcal{I}$ and $\mathcal{A}^\mathcal{F}$ proposed by our paradigm effectively expand the perturbation space, thereby boosting the utilization of unlabeled samples.

\begin{table}[t]
    \scriptsize
    \centering
    \tabcolsep=0.05cm
    \caption{Running speed~(sec/iter) analysis for different configurations of the proposed paradigm. We use FixMatch's threshold in the second student branch, \ie the IFMatch~(Fix) model.}
    \begin{tabular}{c|c|c|c|c}
    \toprule
    $\mathcal{A}^{\mathcal{F}_s}$ in Branch I & $\mathcal{A}^{\mathcal{F}_w}$ in Branch II & CIFAR-10-40 & CIFAR-100-400 & ImageNet-100k \\
    \midrule
     & & 0.1011 & 0.2058 & 0.3914 \\
     & \checkmark & 0.1029 & 0.2092 & 0.3983 \\
    \checkmark & & 0.1114 & 0.2596 & 0.4357 \\
    \checkmark & \checkmark & 0.1140 & 0.2640 & 0.4423 \\
    \bottomrule
    \end{tabular}
    \label{tab:speed}
\end{table}

\section{Running Speed Analysis}
The proposed paradigm demonstrates substantial performance promotion when compared to the traditional approach. However, the obtained improvement is accompanied by feature-level perturbation and a triple-branch structure, inevitably resulting in a reduction in training speed. Consequently, we conduct a running speed analysis for our paradigm. The evaluated training speed is provided in \cref{tab:speed}. The baseline model, as presented in line~1, follows the old paradigm. Moreover, the adaptive weak feature-level perturbation $\mathcal{A}^{\mathcal{F}_w}$~(line~2) in the second student branch introduces negligible costs, indicating the efficiency of implementing feature-level perturbation. Additionally, the first student branch~(line~3 and line~4) leverages strong feature-level perturbation $\mathcal{A}^{\mathcal{F}_s}$ to comprehensively explore the feature perturbation space, albeit incurring non-trivial costs. Overall, the proposed paradigm significantly improves the model's performance at the expense of acceptable additional computational costs.

\begin{table}[t]
    \footnotesize
    \centering
    \tabcolsep=0.1cm
    \caption{Performance on Cityscapes. Feature map translation and shearing~(`movement') are excluded from $\mathcal{A}^\mathcal{F}$ in segmentation tasks. The first line provides the proportion of labeled samples to the total number of available samples.}
    \begin{tabular}{c|c|c|c|c}
    \toprule
    Method~(ResNet-50) & 1/16 & 1/8 & 1/4 & 1/2 \\
    \midrule
    U$^2$PL & 70.6 & 73.0 & 76.3 & 77.2 \\
    UniMatch & 75.0 & 76.8 & 77.5 & 78.6 \\
    IFMatch~(Fix) & \textbf{76.3} & \textbf{77.5} & \textbf{78.3} & \textbf{79.0} \\
    \bottomrule
    \end{tabular}
    \label{tab:ss_seg}
\end{table}

\section{Performance on Semi-Supervised Semantic Segmentation}
In addition to semi-supervised classification, we also test the performance of the proposed paradigm on the semi-supervised semantic segmentation task. As shown in \cref{tab:ss_seg}, IFMatch~(Fix) consistently outperforms U$^2$PL and UniMatch with varying numbers of labeled samples. The impressive performance on different tasks demonstrates the effectiveness and generalization capability of our approach.

\begin{table*}[t]
    \scriptsize
    \centering
    \tabcolsep=0.05cm
    \caption{Detailed training settings for balanced SSL.}
    \begin{tabular}{c|c c c c c}
    \toprule
    Datasets & CIFAR-10 & CIFAR-100 & SVHN & STL-10 & ImageNet \\
    \midrule
    \midrule
    Backbone & WRN-28-2 & WRN-28-8 & WRN-28-2 & WRN-37-2 & ResNet-50\\
    \midrule
    Weight Decay & 5e-4 & 1e-3 & 5e-4 & 5e-4 & 3e-4\\
    \midrule
    $B_L$ / $B_U$ & \multicolumn{4}{c}{64 / 448} & 128 / 128 \\
    \midrule
    Initial Learning Rate & \multicolumn{5}{c}{0.03} \\
    \midrule
    Learning Rate Scheduler & \multicolumn{5}{c}{$\eta=\eta_0 cos(\frac{7 \pi k}{16K})$} \\
    \midrule
    SGD Momentum & \multicolumn{5}{c}{0.9} \\
    \midrule
    Model EMA & \multicolumn{5}{c}{0.999} \\ 
    \midrule
    $\lambda_u$ & \multicolumn{5}{c}{1.0} \\
    \midrule
    $\mathcal{A}^{\mathcal{I}_w}$ / $\mathcal{A}^{\mathcal{I}_s}$ & \multicolumn{5}{c}{Random Crop, Random Horizontal Flipping / RandAug}\\ 
    \midrule
    $\tau$ & \multicolumn{5}{c}{0.95} \\
    \bottomrule
    \end{tabular}
    \label{tab:param_balanced_ssl}
\end{table*}

\begin{table*}[t]
    \scriptsize
    \centering
    \tabcolsep=0.05cm
    \caption{Detailed training settings for imbalanced SSL.}
    \begin{tabular}{c|c c}
    \toprule
    Datasets & CIFAR-10-LT & CIFAR-100-LT\\
    \midrule
    \midrule
    Backbone & \multicolumn{2}{c}{WRN-28-2} \\
    \midrule
    Weight Decay & \multicolumn{2}{c}{4e-5} \\
    \midrule
    $B_L$ / $B_U$ & \multicolumn{2}{c}{64 / 128}\\
    \midrule
    Initial Learning Rate & \multicolumn{2}{c}{2e-3} \\
    \midrule
    Learning Rate Scheduler & \multicolumn{2}{c}{$\eta=\eta_0 cos(\frac{7 \pi k}{16K})$} \\
    \midrule
    Optimizer & \multicolumn{2}{c}{Adam} \\
    \midrule
    Model EMA & \multicolumn{2}{c}{0.999} \\ 
    \midrule
    $\lambda_u$ & \multicolumn{2}{c}{1.0} \\
    \midrule
    $\mathcal{A}^{\mathcal{I}_w}$ / $\mathcal{A}^{\mathcal{I}_s}$ & \multicolumn{2}{c}{Random Crop, Random Horizontal Flipping / RandAug} \\
    \midrule
    $\tau$ & \multicolumn{2}{c}{0.95} \\
    \bottomrule
    \end{tabular}
    \label{tab:param_imbalanced_ssl}
\end{table*}

\section{Implementation Details}
The detailed training settings for balanced and imbalanced semi-supervised learning are presented in Table~\ref{tab:param_balanced_ssl} and Table~\ref{tab:param_imbalanced_ssl}, respectively.

\end{document}